\documentclass[conference,compsoc]{IEEEtran}
\ifCLASSOPTIONcompsoc
  \usepackage[nocompress]{cite}
\else
  \usepackage{cite}
\fi
\ifCLASSINFOpdf
\else
\fi
\usepackage{graphicx}

\usepackage{caption}
\usepackage[left=1.3cm,top=2cm,right=1.3cm,bottom=2.5cm,bindingoffset=0.13in]{geometry}
\usepackage{booktabs,siunitx}

\begin{document}
%
\title{Rotation Adaptive Visual Object Tracking with Motion Consistency}


\author{\IEEEauthorblockN{
		Litu Rout\IEEEauthorrefmark{1}, 
		Sidhartha\IEEEauthorrefmark{2},
		Gorthi R.K.S.S. Manyam \IEEEauthorrefmark{3},
		Deepak Mishra\IEEEauthorrefmark{4}}
	\IEEEauthorblockA{Department of Avionics\\
		Indian Institute of Space Science and Technology\\
		Thiruvananthapuram, Kerala\\
		India - 695 547\\
		Email:\IEEEauthorrefmark{1}liturout1997@gmail.com, 		\IEEEauthorrefmark{2}siddharthaiist@gmail.com, 
		\IEEEauthorrefmark{3}saisubrahmanyam.gorthi@gmail.com, 
		\IEEEauthorrefmark{4}deepak.mishra@iist.ac.in}}


%


\maketitle

\begin{abstract}
	Visual Object tracking research has undergone significant improvement in the past few years. The emergence of tracking by detection approach in tracking paradigm has been quite successful in many ways. Recently, deep convolutional neural networks have been extensively used in most successful trackers. Yet, the standard approach has been based on correlation or feature selection with minimal consideration given to motion consistency. Thus, there is still a need to capture various physical constraints through motion consistency which will improve accuracy, robustness and more importantly rotation adaptiveness. Therefore, one of the major aspects of this paper is to investigate the outcome of rotation adaptiveness in visual object tracking. Among other key contributions, the paper also includes various consistencies that turn out to be extremely effective in numerous challenging sequences than the current state-of-the-art.
\end{abstract}


%
\IEEEpeerreviewmaketitle

\section{Introduction}
	Estimating similarity across various image patches is one of the  most fundamental components in the field of computer vision and pattern recognition. In most of the cases, a precise similarity measure leads a solid foundation for several challenging tasks which include structure from motion, wide baseline matching, object recognition, segmentation, classification and image retrieval \cite{instance}. Among visual object tracking applications, video surveillance, human computer interaction and robotics have been prime focus of computer vision community for many years. Although great achievements have been accomplished within the last few years, there is still a need to optimize the performance in terms of accuracy, robustness and speed.

Over the years, many tracking algorithms have been proposed \cite{vot2016} where the main goal is to localize an object in a series of frames with the sole supervision of a bounding box given only in the first frame of the sequence. In these scenarios the object's appearance model is learnt from previous frame of a sequence online and these learnt models are then cross-correlated with the search image to localize the target object in the next frame. There are several state-of-the-art methods based on similarity rationale with certain modifications such as KCF \cite{kcf}, SRDCF \cite{srdcf} and C-COT \cite{ccot} which are widely accepted by the tracking community. Due to the tremendous achievements of deep neural networks in diverse computer vision applications, the researchers have focused their attention to bring the best out of deep convolutional nets in tracking paradigm. Moreover, the scarcity of large sets of supervised data and extremely slow learning ability have made deep trackers not feasible for real life applications \cite{luca}. Even though there is a massive constraint of speed, deep trackers such as DeepSRDCF \cite{deepS}, TCNN \cite{tcnn} and MDNet \cite{mdnet} have proved their effectiveness in wide variety of challenging sequences.
Some of the recent trackers have aimed to learn a detector per video by fine-tuning multiple layers of a pre-trained deep network with stochastic gradient descent mechanism \cite{mdnet,fc}. But, the necessity of high frames per second in real world applications leaves the online adaptive deep convolutional networks a step behind the other state-of-the-art trackers \cite{vot2016,otb2015}. A possible solution to these shortcomings could be to use a pre-trained deep similarity network such as Siamese network \cite{oneshot} in order to discriminate the target from its background. So the objective of the network would be to learn from a single exemplar image in one branch and predict the essential parameters of the other branch which will assist in identifying instances of the same object in the upcoming frames \cite{oneshot}. Thus, the deep network would generalize a similarity function from annotated pairs of raw image patches without attempting to use hand crafted features \cite{patch}.

Most of the trackers have achieved appealing results both in accuracy and robustness. It is also witnessed that the use of several consistency techniques such as scale adaptive KCF \cite{scaleKCF}, influence of windowing \cite{luca}, bounding box regression \cite{mdnet} and online adaptation of appearance model \cite{mdnet} has turned out to be extremely effective in numerous tough sequences. Even though the introduction of immunity to minor scale changes has brought radical advancement, still there is necessity of orienting the bounding box according to the target object. In real life applications, detecting not only proper bounding boxes but also estimating the orientation of the target object plays a vital role. This would be a key factor in increasing overlap ratio and anticipating trajectory of the target object in a more efficient manner. One of the major advantages of using orientation would be to update the appearance model and to use more sophisticated features from the rotated version of cropped exemplar image in the subsequent frames.

Therefore, one of the important aspects of this paper is to propose a scheme to determine the orientation of the target object and analyse its impact in visual object tracking. We further extend our work and propose a better approach of updating target position taking into account the distance and direction of motion and Gaussian weighted average response map based on various scales. Our proposed method is generic in a sense that it can be applied to any state-of-the-art trackers which inturn improves their performance standard. To establish this proposition, we have integrated our algorithm with SiameseFC \cite{luca} and CFnet \cite{cfnet}. The main reason behind choosing these two trackers is that they achieve comparable state-of-the-art performance while operating at extremely high frames per second. The proposed algorithm have been evaluated on popular tracking benchmarks such as VOT \cite{vot2016} and OTB \cite{otb2015}.


\section{Related Works}
	Most of the correlation filter based algorithms are based on two key elements, how the target object is represented and how to localize this object of interest in the subsequent frames. Object representation models have developed gradually from histogram \cite{kernel} based approach to more advanced generative \cite{gen,gen2} or discriminative \cite{kcf,discr} approaches. For target object localization, methods such as Elliptical head tracking \cite{loc1}, Probabilistic color and adaptive multi-feature tracking \cite{loc2}, Robust visual tracking \cite{loc3} and Learning to track with multiple observers \cite{loc4} have gained a lot of attention. Recently, the widespread success of object detection algorithm has emerged an advanced approach of localization, known as tracking-by-detection. Due to outstanding performance of these tracking-by-detection algorithms on evaluation benchmarks \cite{vot2016,otb2015}, this paradigm has gained popularity in the tracking community. This method usually employ binary classifier to discriminate target object from its background. S. Hare \textit{et al.} discuss Struck \cite{struck}, a discriminative tracker which employs a kernelized structured output Support Vector Machine(SVM) to provide adaptive tracking. M. Danelljan \textit{et al.} have proposed SRDCF \cite{srdcf} which uses a spatially regularized correlation filter that helps in learning from a large set of negative samples, without corrupting the positive samples. Y. Li \textit{et al.} have proposed a scale adaptive scheme in \cite{scaleKCF} which has strong impact on determining the size of the target efficiently. G. Koch \textit{et al.} explores a method to train Siamese neural network which ranks the similarity between its input image patches \cite{siameseOneshot}.  J. Valmadre \textit{et al.} take a step forward to investigate the influence of modified Alex-net on Siamese network \cite{luca} and propose an end-end training model of correlation filter in CFnet \cite{cfnet}. 

\begin{figure}[t]
	\begin{center}
		\includegraphics[width=0.95\linewidth]{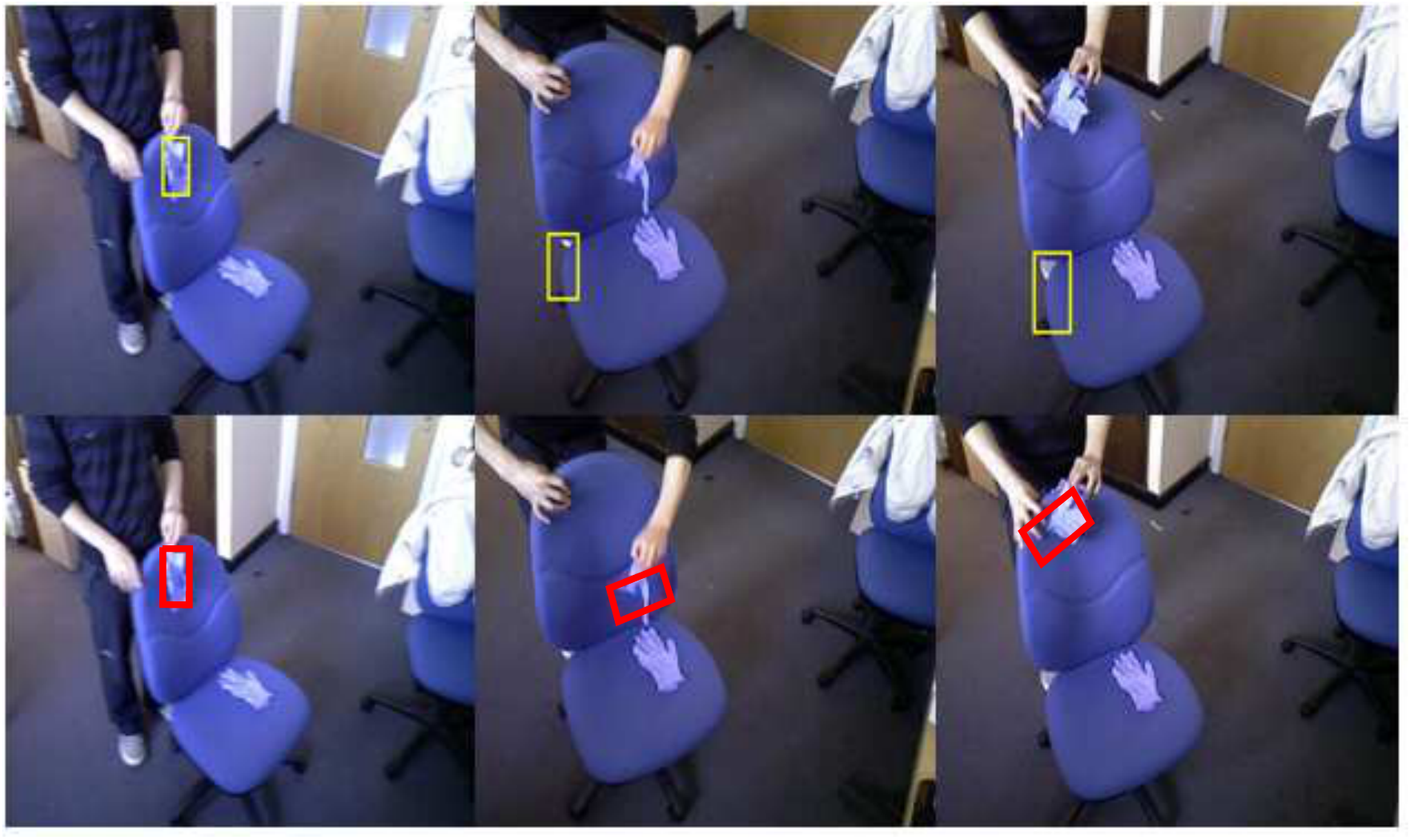}
	\end{center}
	\caption{ Sample frames from glove sequence regarded as one of the toughest sequences according to VOT 2016 results \protect\cite{vot2016}. First column indicates the ground truth bounding box in the first frame. Our modified SiameseFC(red) successfully tracks the geometric deformations unlike original SiameseFC\protect\cite{luca}(yellow).}
	\label{glove}
\end{figure}

Sequence like glove \cite{vot2016} from VOT challenge as shown in Figure \ref{glove} undergoes severe deformation which leads to significant changes in aspect ratio as well as rotation of the bounding box. Due to this, the ill-equipped axis aligned bounding box used in most state-of-the-art tracking methods fails to capture more detailed information from the object of interest. Y. Hua \textit{et al.} \cite{yanghua} have addressed this issue by generating suitable candidates which capture more detailed information by estimating the transformations undergone by the object. H. A. Rowley \textit{et al.} have carried out an extensive research in designing a rotation invariant neural network based face detection system \cite{rotFace}. There are several face detection systems which could only detect upright or frontal faces \cite{rotface1}. But this is far from the reality where images of faces could be much more complicated than just upright or frontal. Introduction of router network in \cite{rotFace} to detect angle of orientation of the face has helped to convert the rotated faces back to frontal before it is actually passed through the face detector. M. Jaderberg \textit{et al.} have introduced a spatial transformer network (STN) \cite{stn}, where the model learns invariance to rotation, scale, translation and other warping of the input data. Use of localization network, sampling grid generator and sampler in a STN have brought radical achievement in challenging object classification datasets such as CUB-200-2011 birds dataset \cite{stnD}. In this paper we explore extensively the impact of rotation invariance in tracking paradigm as well as several new consistency techniques which have out performed the original tracking algorithm by a large margin. In the following section \ref{method} and section \ref{exp} , we explain the detailed modifications and experimental results based on our evaluation on OTB \cite{otb2015} and VOT \cite{vot2016} datasets respectively.
	\section{Proposed Methodology}\label{method}
In this section, we detail our contributions  by proposing a generic approach for incorporating rotation invariance (RI) object tracking and introducing the motion consistencies guided by the laws governing physical motion of the objects. For the sake of experimentation and analysis we incorporated the proposed modifications to Siamese Net and CFnet. Here we first briefly discuss about the architecture of SiameseFC in \ref{siamese} and CFnet in \ref{cfnet} followed by  the consistencies termed as Displacement Consistency in \ref{dv}, Scale Consistency in \ref{s} and RI in \ref{r}. 

\subsection{Siamese Fully Convolutional Network} \label{siamese}
\begin{figure}[t]
	\centering
	\includegraphics[width = 0.8\columnwidth]{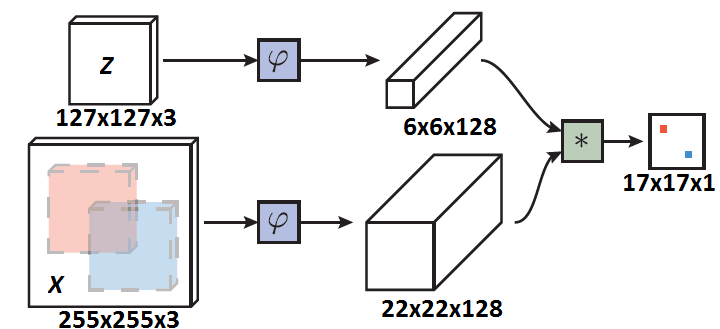}
	\caption{Siamese Fully Convolutional Network. $z$-branch and $X$-branch are known as exemplar and instance branch respectively. $Z$ is the target exemplar image extracted from the first frame. Siamese network learns from the sole supervision of $Z$ which is done once per sequence. $X$ is the instance image. Output response map predicts the possible target location based on the highest response score \protect\cite{luca}.}
	\label{siameseFC}
\end{figure}

Deep similarity learner mainly learns the parameters of a function $f(z,x) $ which takes two images as input and generates a response score. If the two images depict the same object, it generates a high score otherwise a low score. A convolutional neural net is used as this learning function $f(z,x) $. Typically Siamese architectures have been quite successful in deep similarity measure. The network learns the parameters from the first frame of each sequence and then all the possible candidates are tested exhaustively to measure similarity with the exemplar image. A simple architecture of Siamese fully convolutional network is shown in Figure \ref{siameseFC}. In this architecture, the appearance model of the exemplar image isn't updated at all. Siamese network employs same transformation $\phi$ which is a five layered CNN to both of its inputs $z$ and $x$. The transformed inputs $\phi(z)$ and $\phi(x)$ are cross-correlated to obtain a response map. The location of maximum response score indicates the position of the object of interest. Thus, the overall similarity function becomes $f(z,x) = g(\phi(z),\phi(x))$. One of the major advantages of using fully convolutional architecture is that, a larger search region can be fed into the network without resizing it to the size of exemplar. Then the similarity function helps in finding the response score at all translated sub-windows in a large search region. The response map computed by this architecture is given by equation \ref{fzx}.
\begin{equation}\label{fzx}
	f(z,x) = \phi(z) * \phi(x) + b
\end{equation}
where b $\epsilon$ $\mathbf{R}$ is a bias signal.

A very detailed description of Siamese architecture along with mathematical discussion can be found in \cite{luca,instance}.
\subsection{Correlation Filter Network} \label{cfnet}
\begin{figure}
	\centering
	\includegraphics[width = 0.8\columnwidth]{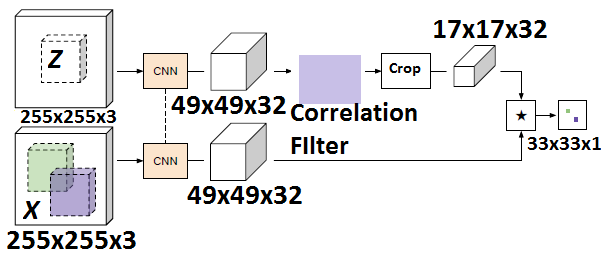}
	\caption{Correlation filter network. An additional correlation filter block in exemplar branch brings robustness to translation. End-to-end training model of a correlation filter which uses CNN features \protect\cite{cfnet}.}
	\label{cfnetFig}
\end{figure}

Correlation filter network is a modification over baseline Siamese network as given in equation \ref{fzx}. The correlation filter network is shown in Figure \ref{cfnetFig}. The new similarity measure function is given in equation \ref{cfnetE}.
\begin{equation} \label{cfnetE}
	h(z,x) = sW(\phi(z)) * \phi(x) + b
\end{equation}
where $s$ and $b$ are scale and bias respectively. The correlation filter block $W(x)$ uses feature map $\phi(z)$ to learn a template by solving ridge regression problem in the Fourier domain \cite{kcf}. The effect of circular boundaries has been mitigated by multiplying $\phi(z)$ with a cosine window and cropping the final template. Unlike SiameseFC, the appearance model in CFnet is updated after each frame using a rolling average in order to avoid abrupt transitions from frame to frame. The rest of the procedure is similar to that of the baseline Siamese as illustrated in section \ref{siamese}. A deep insight into CFnet including back propagation can be found in \cite{cfnet}.

\subsection{Displacement Consistency}\label{dv}
\begin{figure}
	\centering
	\includegraphics[width = 0.6\columnwidth]{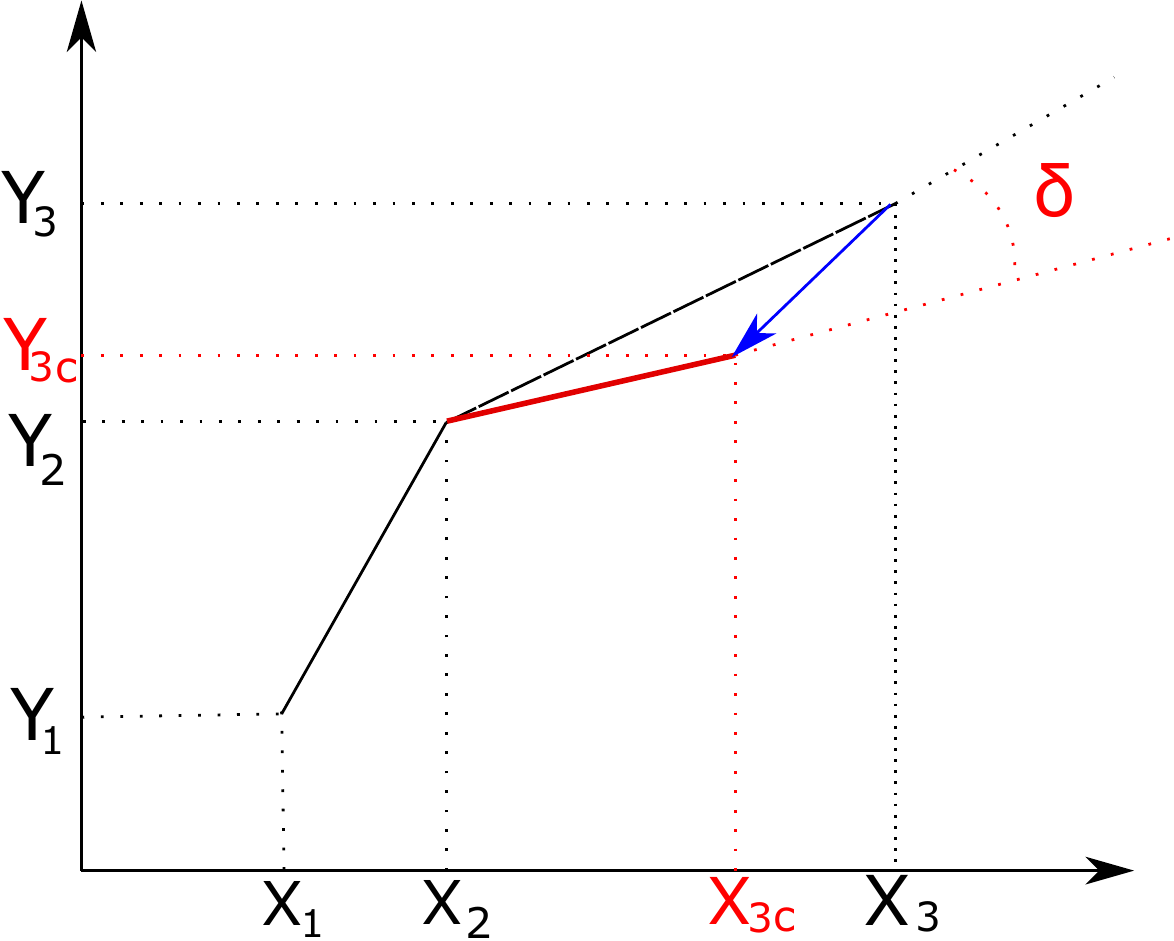}
	\caption{Conventional target centroid update strategy. Let $[X_{1},Y_{1}]$ and $[X_{2},Y_{2}]$ represent the target centroids in the first and second frame respectively. Let $ [X_{3},Y_{3}]$ represent the predicted centroid in the third frame. Let $[X_{3c},Y_{3c}]$ represents the updated centroid in the third frame. Let $\delta$ represents the angular deviation occurred due to conventional centroid update.}
	\label{tPos}
\end{figure}

\begin{figure}
	\centering
	\includegraphics[width = 0.7\columnwidth]{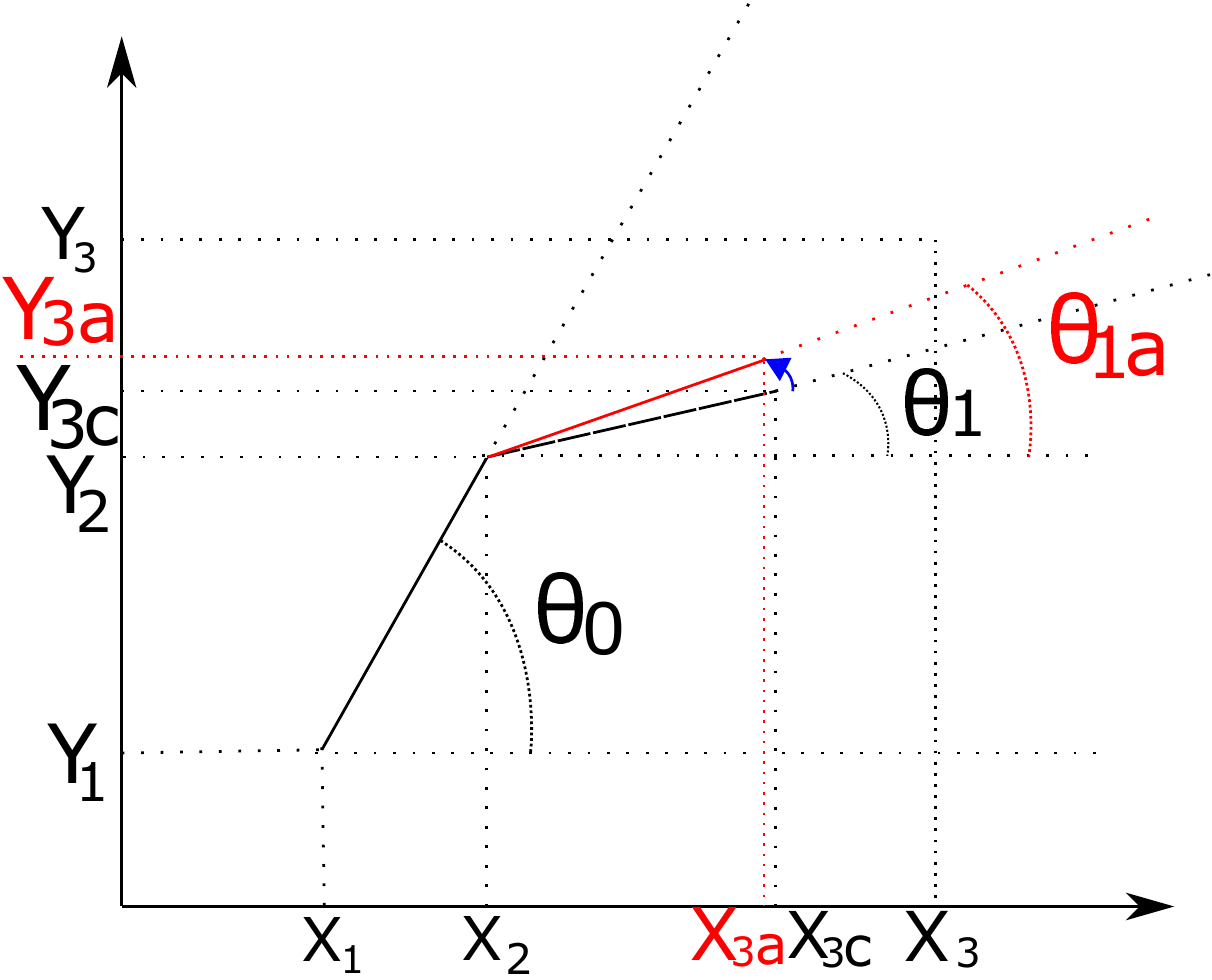}
	\caption{Angle consistency. Let $\theta_{1}$ represents the angle of the centroid in the third frame with respect to the centroid in the second frame. Let $\theta_{0} $ represents the angle of the centroid in the second frame with respect to the first. Let $\theta_{1n}$ represents the updated angle in the third frame. Let $[X_{3a},Y_{3a}]$ represents the new updated centroid by using equation \ref{angCorrE} with 1\% weight given to previous angle i.e. $w_{\theta}$ = 0.01. }
	\label{angCorr}
\end{figure}

\begin{figure}
	\centering
	\includegraphics[width = 0.7\columnwidth]{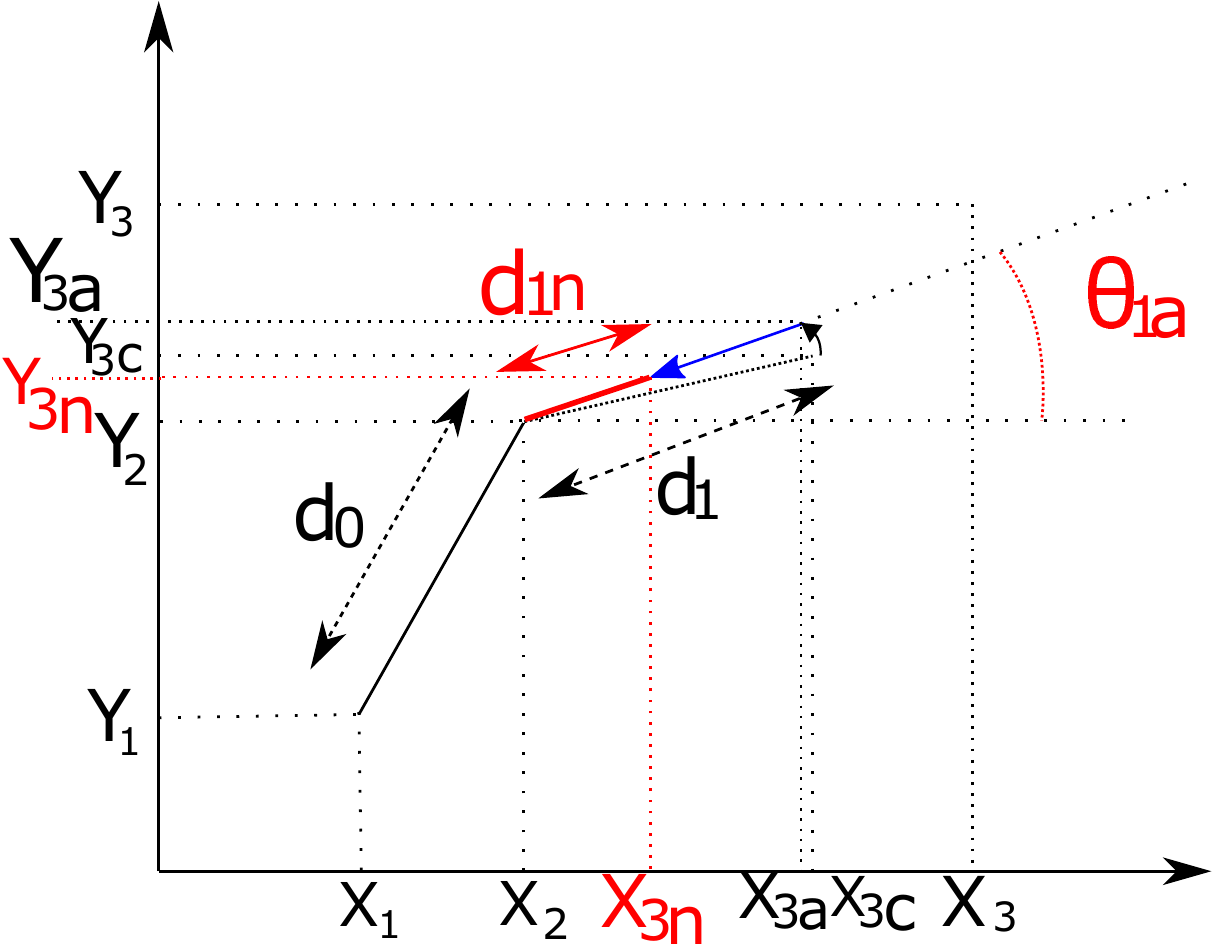}
	\caption{Distance consistency. Let $d_{0}$ represents the distance of centroid from frame 1 to 2. Let $d_{1}$ represents the distance from frame 2 to 3 after angle consistency. Let $d_{1n}$ represents the updated distance obtained by using equation \protect\ref{dCorrE} with 1\% preference given to previous distance i.e. $w_{d}$ = 0.01. Let $[X_{3n},Y_{3n}]$ represents the final position of the centroid after Displacement consistency.   }
	\label{dCorrF}  
\end{figure}

In order to avoid the target position deviating much from its previous position, most trackers \cite{luca,cfnet} employ a target centroid update strategy as shown in Figure \ref{tPos}. Mathematically, this update strategy can be written as  in equation \ref{tPosE}. This is usually employed to enforce smoothness on the object motion. However, we identify that this is not sufficient to enforce the smoothness on the displacement (direction and distance) of the object in motion which can contribute to more reliable tracking. This can be noticed by observing the Figure \ref{bird1}. Due to the incremental angular deviation $\delta$, the target centroid keeps drifting away from the actual centroid which decreases the overlap ratio. It is observed from Table \ref{dvsrT} that minimizing this deviation $\delta$ as proposed has increased the success as well as precision.
\begin{equation}\label{tPosE}
	[X_{3c},Y_{3c}] = w \times [X_{2},Y_{2}] + (1-w) \times [X_{3},Y_{3}] 
\end{equation}
We have integrated a new displacement consistency approach on top of the conventional approach to enhance the degree of smoothness. The angle consistency is illustrated in Figure \ref{angCorr}.  In distance consistency, the algorithm remembers the previous distance encountered and updates the new distance using rolling average. A pictorial representation of distance consistency is elucidated in Figure \ref{dCorrF}. After displacement consistency, the new centroid of the target is computed using equation \ref{daCorrE}. The accuracy and robustness after these integrations along with original values have been provided in section \ref{exp} which proves the efficacy of the scheme. 
\begin{equation}\label{angCorrE}
	\theta_{1n} = w_{\theta} \times \theta_{0} + (1-w_{\theta}) \times \theta_{1} 
\end{equation}
\begin{equation}\label{dCorrE}
	d_{1n} = w_{d} \times d_{0} + (1-w_{d}) \times d_{1}
\end{equation}
\begin{equation}\label{daCorrE}
	[X_{3n},Y_{3n}] = [X_{2},Y_{2}] + d_{1n}\angle{\theta_{1n}} 
\end{equation}
\subsection{Scale Consistency}\label{s}
The conventional approach to estimate size of the target object is to form a scale pyramid and compute response map using each of these images \cite{scaleKCF}. The corresponding scale of the response map having maximum response score among all these response maps determines the size of the target object. Then that particular response map is used for obtaining the target centroid. In this standard approach, only the winning response map i.e the map having maximum response score among all maps decides the size of the object. However, as we know that in real scenarios,  the scale of the object doesn't undergo drastic change from frame to frame as the scale change depends on the distance of the object from camera and as the objects move smoothly in many real scenarios. Though there are methods which add a penalty factor to the new target size, this applies only to the size of the target object. However, if the position of the target centroid itself is corrupted due to the use of wining response map only, it will persist in subsequent frames. In this standard scenario, the response maps that correspond to different scales aren't used in determining the centroid. Therefore, we propose to use Gaussian weighted average response map centred at the winning map and  have variance as an additional hyper parameter. In this way we can incorporate the response maps that correspond to various scales in the scale pyramid. Our approach has enhanced the accuracy as well as robustness of the considered base trackers. The results of this experiment are described in details in section \ref{exp}. The pseudo code for Gaussian weighted average response map is provided in Algorithm I.\\
\rule{\columnwidth}{2pt}	\\
Algorithm I : Scale Consistency using Gaussian weights\\
\rule{\columnwidth}{1pt}\\
\begin{enumerate}
	\item \textbf{Input parameters :}
	
	Let \textit{responseMaps} represents the stack of response maps at each scale. $\mu$ represents the index of the winning response map. $\sigma_{scale}$ represents the standard deviation of Gaussian weights. \textit{scaleBins} numerically represents each scale i.e. \textit{scaleBins(1)} represents the first scale, \textit{scaleBins(2)} represents the second scale and so on. Let $N$ represents the total number of scales used in the scale pyramid.
	
	\item \textbf{Computation of scale weights and updation of \textit{responseMap}:}
	\begin{enumerate}
		\item Define weights for each scale as\\ \textit{scaleWeights} = $\frac{1}{\sqrt{2 \times \pi} \times \sigma_{scale}} \exp^{- (\frac{scaleBins - \mu}{\sigma_{scale}})^2}$
		\item $responseMap =\\
		\sum_{i = 1}^{N} [responseMaps(i) \times scaleWeights(i)]$
	\end{enumerate}
	\item \textbf{Output response map :} The output of this algorithm is the Gaussian weighted average \textit{responseMap}.
\end{enumerate}	
\rule{\columnwidth}{2pt}	

\subsection{Rotation Invariance}\label{r}
In this section \ref{r}, we will discuss two different ways of incorporating rotation adaptiveness in tracking algorithms such as the proposed rotation invariant SiameseFC \ref{rotSiam} and rotation invariant CFnet \ref{rotCF}. The former can be used where the target object is not updated after each frame and the later can be used where the object is updated after each frame.
\begin{figure*}
	\centering
	\includegraphics[width = 0.7\textwidth]{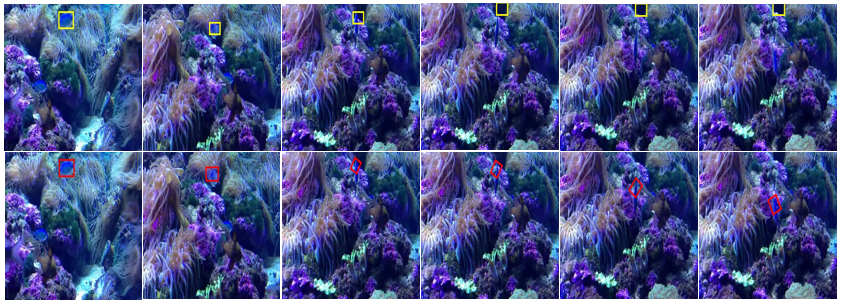}
	\caption{Sample frames from fish1 sequence denoted as one of the toughest sequences according to VOT 2016 results \protect\cite{vot2016}. First column indicates the ground truth bounding box in the first frame. Our modified SiameseFC(red) successfully tracks the geometric deformations unlike original SiameseFC\protect\cite{luca}(yellow).}
	\label{fish1}  
\end{figure*}
\subsubsection{Rotaion Invariant SiameseFC}\label{rotSiam}
\begin{figure*}[t]
	\centering
	\includegraphics[width = 0.7\textwidth]{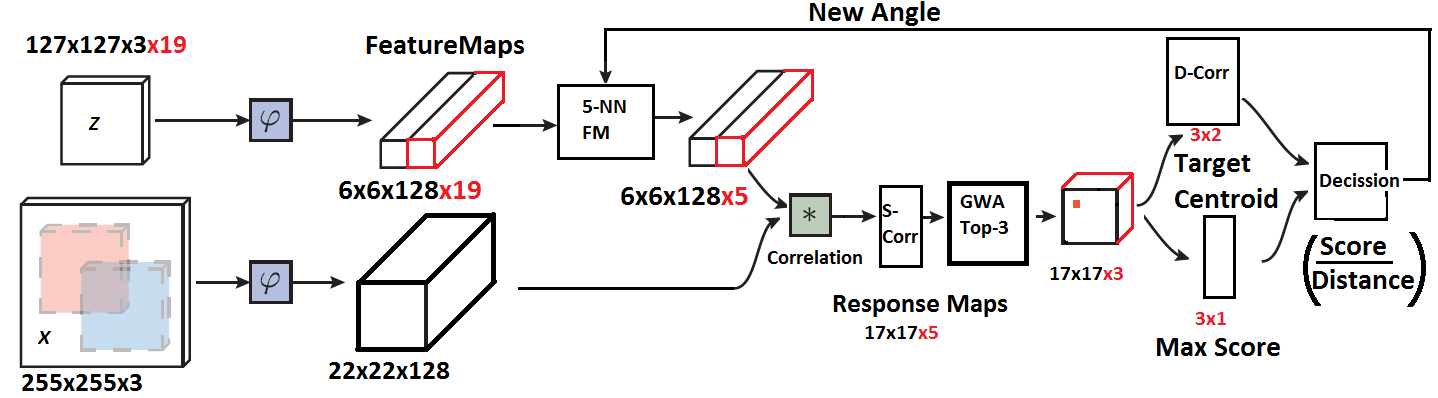}
	\caption{Rotation Invariant Siamese Fully Convolutional Network. The conventional SiameseFC extracts features for cropped exemplar, instead exemplar image can be rotated uniformly from -180$^{\circ}$ to 180$^{\circ}$ at an interval of $\theta$ and corresponding features can be extracted. Here, $\theta$ = 20$^{\circ}$. So there would be 19 feature maps instead of only one. Assume initial newAngle to be 0$^{\circ}$. 5-NN FM block passes 5 nearest neighbour feature maps based on the new angle of rotation. Let S-Corr and D-Corr blocks represent scale \protect\ref{s} and displacement \ref{dv} corrections respectively. GWA block computes Gaussian weighted average response map centred at top 3 maximum score response maps \protect\ref{s}. Decision block computes the ratio of maximum score i.e. probability of detection and the corresponding distance from the previous location. The maximum ratio determines the final target centroid and the new angle of rotation which is determined as the angle corresponding to maximum ratio.}
	\label{modSiamF}
\end{figure*}

When an object is in motion it can assume any of its rotated form or view from frame to frame. However, conventionally only a base template with a fixed (zero) orientation is employed to find the similarity. To incorporate robust RI tracking, we propose to augment various possible rotated images of the object and measure similarity with all these rotated images. The working of rotation invariant Siamese fully convolutional network has been illustrated in Figure \ref{modSiamF}. Since the appearance model isn't updated during tracking, the corresponding features of rotated exemplar can be extracted once for a sequence. Since the angle of rotation does not change drastically from frame to frame, only 5 nearest neighbour response maps are used in computing response map. The mean of the Gaussian weights is considered as the index of the winning response map and variance is tuned as an additional hyper-parameter in the similar manner as explained for different scales in \ref{s}. In order to avoid false alarm, we have computed three Gaussian weighted average response maps centred at top three maps according to their response scores. Accordingly there would be three most probable target centroids, out of which the final centroid is selected based on the highest score to displacement ratio in a sense that the object wouldn't have travelled far from its previous location. In this approach, as the path with dominant direction is detected, the bounding box can be rotated accordingly to increase the overlap ratio. A comparison between SiameseFC and rotation invariant SiameseFC is shown in Figure \ref{fish1}. We have evaluated our rotation invariant SiameseFC on VOT datasets \cite{vot2016} and the obtained results are provided in \ref{exp}.

\subsubsection{Rotation Invariant CFNet}\label{rotCF}
\begin{figure*}[t]
	\centering
	\includegraphics[width = 0.7\textwidth]{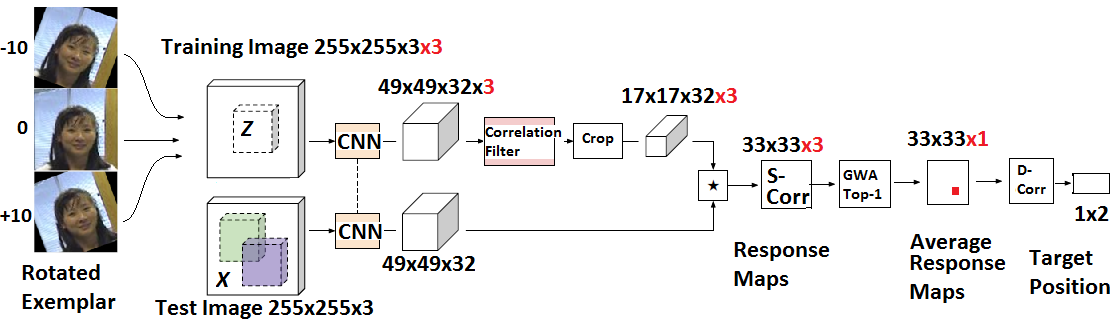}
	\caption{Rotation Invariant Correlation filter network. The input exemplar image is rotated by $\theta$ = [-$\zeta$, 0$^{\circ}$, +$\zeta$]. Here, $\zeta$ = 10$^{\circ}$ represents the angle of rotation of the exemplar. The angle of rotation 0$^{\circ}$ represents the actual cropped exemplar image obtained after each iteration. Thus, the three feature maps of rotated exemplar are correlated with the feature map of instance image which produce three most probable response maps. Let S-Corr and D-Corr blocks represent scale \protect\ref{s} and displacement \protect\ref{dv} corrections respectively. The S-Corr block performs scale correction on these three response maps. The GWA block computes Gaussian weighted average response map centred at the winning response map. The D-Corr block performs displacement correction and computes the final target centroid.}
	\label{modCFFig}
\end{figure*}

Unlike rotation invariant SiameseFC, in rotation invariant CFnet the exemplar is updated after every frame \cite{cfnet}.  In the second frame the object itself would have undergone some rotation. So there is no need to extract features from all the rotated exemplars beforehand, instead only forward and backward rotations after each model update would suffice. Therefore, there is no need to feed back the angle of  rotation. Thus, there would be three response maps corresponding to each of the three rotations. Since we have only three response maps corresponding to three nearest neighbour angles and each response map itself is a Gaussian weighted average map performed by the S-Corr block, there is no need to compute average of the three rotated maps again. The final centroid is computed by using the map having highest response score among all the three. In fact, using Gaussian weighted average after scale correction doesn't seem to improve the performance much, though it would be useful when more nearest neighbour rotated exemplars are used.  This is a general approach which can be integrated into any state-of-the-art trackers to enhance their performance further. The rotation invariant CFnet is illustrated in Figure \ref{modCFFig}. The results of our proposed CFnet DS and CFnet DSR can be found in section \ref{exp}. In rest of the sections, we have referred CFnet-conv2 \cite{cfnet} as original CFnet and applied our consistency strategy to this model. We have used CFnet-conv2 in our experiments because it has less than 4\% parameters used in five-layer baseline and outperforms the rest in the series\cite{cfnet}.
\begin{figure*}[t]
	\centering
	\includegraphics[width = 0.7\textwidth]{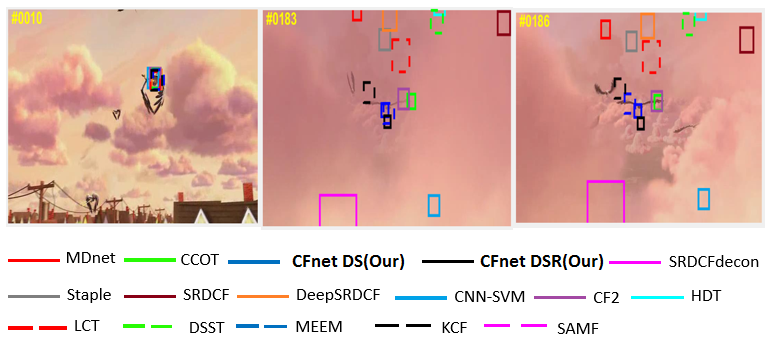}
	\caption{Sample frames from Bird1 sequence, one of the toughest sequences  in OTB50 \protect\cite{otb2015}. The results are obtained using fully integrated OTB toolkit. Our trackers have not deviated much from the target centroid mainly due to the integration of displacement correction.}
	\label{bird1}  
\end{figure*}
\section{Experimental Details and Analysis}\label{exp}
We have evaluated original CFnet and our modified CFnet on 43 sequences out of OTB50 with 3 repetitions for each sequence in order to get a rough estimation of the performance. These particular sequences were selected based on the toughness of deformations incurred in the object of interest. For the evaluation on 43 sequences, we have used OTB-TRE function which has been provided in original CFnet \cite{cfnet} codes repository. The tracking performance varies from machine to machine based on whether GPU support is enabled. The reason for this is the numerical effects which gets accumulated over time. Due to which, most of the trackers suffer from a slight variation in results when re-evaluated on different machines. In order to avoid this effect, we have evaluated the original tracker and all our modifications under exactly same circumstances i.e. same sequences, same system and same evaluation function. The results are given in Table \ref{dvsrT}. From Table \ref{dvsrT}, it is clearly observed that our proposed displacement \ref{dv} and scale \ref{s} correction schemes have improved the performance of original CFnet \cite{cfnet}. The success and precision values for different angles of rotation are shown in Table \ref{cfnetRot}. From Table \ref{cfnetRot}, it is clear that the tracker achieves optimal performance for the angle of rotation $\zeta$ = 8$^{\circ}$ and its performance decreases with increase in angle of rotation. This is evident from practical point of view as there won't be drastic change in orientation between two subsequent frames.

\begin{table*}[]
	\centering
	\begin{tabular}{|l|l|l|l|l|l|l|}
		\hline
		Angle of Rotation    & 5     & 7     & 8     & 10    & 20    & 30    \\
		\hline\hline
		Success(AUC)         & 56.97 & 58.11 & 58.78 & 57.01 & 53.80 & 48.17 \\
		Precision(threshold) & 67.05 & 68.30 & 69.13 & 67.91 & 67.18 & 63.88\\
		\hline
	\end{tabular}
	\caption{success(AUC) and precision(Threshold) vs various angle of rotation. A rough estimation to obtain optimal angle of rotation $\zeta$ using OTB-TRE function given in CFnet \protect\cite{cfnet} code repository.}
	\label{cfnetRot}
	
\end{table*}

\begin{table}
	\begin{center}
		\begin{tabular}{|l|c|c|}
			\hline
			Tracker  & Success(AUC)                                                            & Precision (Threshold)  \\
			\hline\hline
			
			CFnet                                                            & 57.3447                                                                  & 65.0869  \\
			[0.2cm]
			CFnet D  & 59.0531 & 69.3120 \\[0.2cm]
			CFnet DS  & 59.0531 & 70.3673 \\[0.2cm]			
			CFnet DSR & 58.7865 & 69.1349 \\
			($\zeta$ = 8$^{\circ}$) & & \\
			\hline
		\end{tabular}
	\end{center}
	\caption{Integration of Displacement \protect\ref{dv}, Scale \protect\ref{s} Correction and rotation invariant strategy \protect\ref{r}. DSR stands for Displacement correction, Scale correction and Rotation invariant strategy respectively.The results are obtained using the OTB-TRE evaluation function provided in original CFnet \protect\cite{cfnet} code. This evaluation is similar to OTB toolkit evaluation, but not exact. The reason behind using this evaluation function is to make a fair comparison with the original CFnet \protect\cite{cfnet}. The comparison is done for axis aligned bounding box results.}
	\label{dvsrT}
\end{table}

\begin{figure}
	\centering
	\includegraphics[width = 0.8\columnwidth]{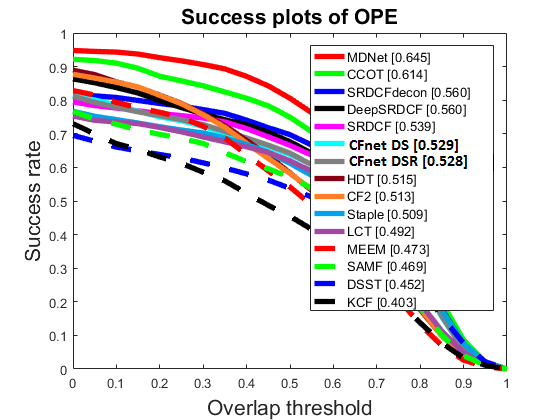}
	\caption{ OTB50 success plot(AUC) obtained using OTB toolkit (Values are scaled down from 100 to 1 by the toolkit). Original CFnet-conv2 success rate(OPE) for OTB50 is equal to 0.527 as per the evaluation in our system. }
	\label{succ}
\end{figure}

\begin{figure}
	\centering
	\includegraphics[width = 0.8\columnwidth]{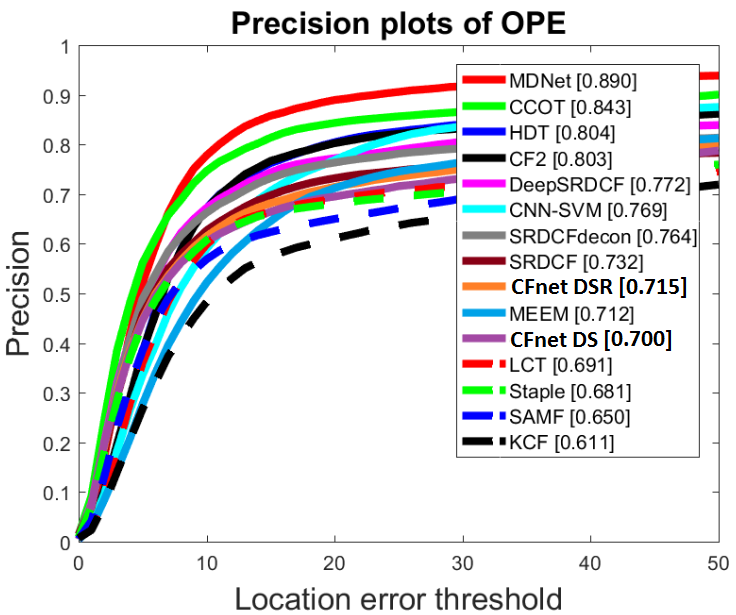}
	\caption{ OTB50 precision plot obtained using OTB toolkit. Original CFnet-conv2 precision(OPE) for OTB50 is equal to 0.702 \protect\cite{cfnet} as per the evaluation in our system. }
	\label{prec}
\end{figure}

Final evaluation is done using OTB toolkit for all 50 sequences \cite{otb2015} and only important results are shown due to space constraint. The success rate of original CFnet could not be plotted because of the unavailability of final bounding boxes in OTB results database during the time of writing the paper. A comparison between our modified CFnet and current state-of-the-art trackers is shown in Figure \ref{bird1}. Due to lack of rotated bounding box results of other trackers, we had to use axis aligned bounding box during accuracy assessment. So there is a slight improvement in accuracy and precision as shown in Table \ref{fevalT}.  This performance will certainly be enhanced if compared with rotated bounding box results which unfortunately isn't available for most of the trackers. The success(AUC) and precision plots obtained by using fully integrated OTB toolkit are shown in Figure \ref{succ} and Figure \ref{prec} respectively. 

\begin{table}
	\begin{center}
		\begin{tabular}{|l|c|c|}
			\hline
			Tracker  & Success(AUC)                                                            & Precision (Threshold)  \\
			\hline\hline
			
			CFnet                                                            & 52.7                                                                  & 70.2  \\
			[0.2cm]
			CFnet DS  & 52.9 & 70.0 \\[0.2cm]
			CFnet DSR & 52.8 & 71.5 \\
			($\zeta$ = 8$^{\circ}$) & & \\
			\hline
		\end{tabular}
	\end{center}
	\caption{Fully integrated OTB-OPE comparison. The results are obtained using OTB toolkit.}
	\label{fevalT}
\end{table}

\begin{figure}
	\centering
	\includegraphics[width =0.5\columnwidth]{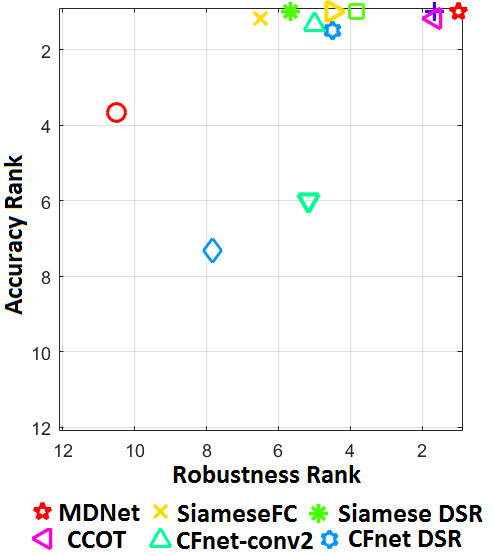}
	\caption{VOT Ranking Plot for experiment baseline(mean). }
	\label{rankVot}
\end{figure}

As per the results obtained using fully integrated vot-toolkit as shown in Table \ref{fVotT}, an improvement of 15.57\% in accuracy rank and 14.3\% in  robustness rank have been observed with no degradation in overlap ratio. Since the absolute value of the accuracy and robustness rank varies based on the trackers used for evaluation, the relative improvements of proposed methods over the original have been used to showcase the efficiency. The ranking plot for experiment baseline is shown in Figure \ref{rankVot}, which depicts the improvement in accuracy and robustness of proposed modified Siamese DSR over the original SiameseFC. 

\begin{table}[]
	\centering
	\begin{tabular}{|l|l|l|ll|}
		\hline
		Tracker     & \begin{tabular}[c]{@{}l@{}}Accuracy\\ Rank\end{tabular} & \begin{tabular}[c]{@{}l@{}}Robustness\\ Rank\end{tabular} & \begin{tabular}[c]{@{}l@{}}Mean\\ Overlap\end{tabular} &  \\
		\hline\hline
		MDNet       & 1.00                                                    & 1.00                                                      & 0.57    &  \\
		CCOT        & 1.17                                                    & 1.67                                                      & 0.54    &  \\
		SiameseFC   & 1.17                                                    & 6.50                                                      & 0.52    &  \\
		Siamese DSR & 1.00                                                    & 5.67                                                      & 0.52    & \\
		CFnet-conv2 & 1.33                                                    & 5.00                                                      & 0.52    & \\
		CFnet DSR & 1.50                                                    & 4.50                                                      & 0.52    & \\
		\hline
	\end{tabular}
	\caption{VOT AR ranking for experiment baseline. Siamese DSR represents the proposed modified network over original SiameseFC\protect\cite{luca}. CFnet DSR represents the proposed modified network over actual CFnet-conv2 \protect\cite{cfnet}}
	\label{fVotT}
\end{table}

\section{Conclusion}
The principal focus of this paper have been to investigate the consequences of rotation adaptiveness in object tracking. The proposed consistency techniques have surely outperformed the original tracking algorithm. The success rate has improved by 4.6\% whereas precision, by 6.75\% as given in the Table \ref{dvsrT}. According to the evaluation of proposed Siamese DSR on VOT \cite{vot2016}, a drastic improvement in robustness rank by 15.7\% and accuracy rank by 14.3\% have been achieved with no degradation in overlap ratio as shown in Table \ref{fVotT}. Above all, detecting the orientation of the target object will certainly be a significant boost in the tracking paradigm. As per the analysis, the concept of rotation adaptive tracking with aforementioned motion consistencies has been exceptional in determining the target centroid in most of the tough sequences in popular tracking benchmarks. As per our investigation of rotation adaptive Siamese and CFnet, we conclude that rotation adaptiveness has certainly enhanced the robustness and regarding the overlap ratio, we suggest that the rotation invariance would surely be more effective if compared with rotated bounding box results. Our future research may include replacing the simple CNN  present in both Siamese and CFnet architectures with a very deep CNN. We will also investigate the improvement achieved by integrating the proposed motion consistency in top trackers such as MDnet \cite{mdnet}, TCNN \cite{tcnn} and CCOT \cite{ccot}.


%
%
%



%

\end{document}